%% file: main.tex
\documentclass{article}
\usepackage[utf8]{inputenc}
\usepackage[letterpaper, margin=1in]{geometry}
\input{preamble.tex}

\newcommand{\exponential}[1]{\text{exp}\left(#1\right)}
\newcommand{\real}{\mathbb{R}}
\newcommand{\C}{\boldsymbol{C}}
\newcommand{\radiance}{\mathbf{c}}
\newcommand{\weight}{\alpha}
\newcommand{\density}{\sigma}
\newcommand{\trans}{\mathcal{T}}
\newcommand{\x}{\mathbf{x}}
\newcommand{\ray}{\mathbf{r}}
\newcommand{\rayOri}{\mathbf{o}}
\newcommand{\rayDir}{\mathbf{d}}

\title{Volume Rendering Digest (for NeRF)}
\author{Andrea Tagliasacchi$^{1,2}$ \quad Ben Mildenhall$^1$
\\[.5em]
$^1$Google Research \quad $^2$Simon Fraser University}
\date{}

\begin{document}
\maketitle

Neural Radiance Fields~\cite{mildenhall2020nerf} employ simple volume rendering as a way to overcome the challenges of differentiating through ray-triangle intersections by leveraging a probabilistic notion of visibility.
This is achieved by assuming the scene is composed by a cloud of light-emitting particles whose density changes in space~(in the terminology of physically-based rendering, this would be described as a volume with absorption and emission but no scattering~\cite[Sec 11.1]{pharr2016physically}.
In what follows, for the sake of exposition simplicity, and without loss of generality, we assume the emitted light \textit{does not} change as a function of view-direction.
This technical report is a condensed version of previous reports~\cite{max2005local, max2010local}, but rewritten in the context of NeRF, and adopting its commonly used notation\footnote{If you are interested in borrowing the LaTeX notation, please refer to:~\url{https://www.overleaf.com/read/fkhpkzxhnyws}}.

\paragraph{Transmittance}
Let the density field $\density(\x)$, with $\x {\in} \real^3$ indicate the differential likelihood of a ray hitting a particle~(i.e. the probability of hitting a particle while travelling an infinitesimal distance).
We reparameterize the density along a given ray $\ray{=}(\rayOri, \rayDir)$
as a scalar function $\density(t)$, since any point $\x$ along the ray can be written as $\ray(t){=}\rayOri{+} t\rayDir$.
Density is closely tied to the transmittance function $\trans(t)$, which indicates the probability of a ray traveling over the interval $[0, t)$ without hitting any particles.
Then the probability $\trans(t {+} dt)$ of \emph{not} hitting a particle when taking a differential step $dt$ is equal to $\trans(t)$, the likelihood of the ray reaching $t$, times $(1 - dt \cdot \density(t))$, the probability of not hitting anything during the step:
\begin{align}
\trans(t+dt) =& \trans(t) \cdot (1 - dt \cdot \density(t))
\\
\frac{\trans(t+dt) - \trans(t)}{dt} \equiv& \trans'(t) = -\trans(t) \cdot \sigma(t) 
\label{eq:derivative}
\end{align}
This is a classical differential equation that can be solved as follows:
\begin{align}
\trans'(t) &= -\trans(t) \cdot \density(t) \\
\frac{\trans'(t)}{\trans(t)} &= -\density(t) \\
\int_a^b \frac{\trans'(t)}{\trans(t)} \; dt &= -\int_a^b \density(t) \; dt \\
\left. \log \trans(t) \right|_a^b &= -\int_a^b \density(t) \; dt \\
\trans(a \rightarrow b) \equiv \frac{\trans(b)}{\trans(a)} &= \exponential{-\int_a^b \density(t) \; dt}   
\label{eq:trans_ab}
\end{align}
where we define $\trans(a \rightarrow b)$ as the probability that the ray travels from distance $a$ to $b$ along the ray without hitting a particle, which is related to the previous notation by~$\trans(t) = \trans(0\rightarrow t)$.

\newpage
\paragraph{Probabilistic interpretation}
Note that we can also interpret the function $1-\trans(t)$ (often called \emph{opacity}) as a cumulative distribution function~(CDF) indicating the probability that the ray \emph{does} hit a particle sometime before reaching distance $t$. Then $\trans(t) \cdot \density(t)$ is the corresponding probability density function (PDF), giving the likelihood that the ray stops precisely at distance~$t$. 

\paragraph{Volume rendering}
We can now calculate the expected value of the light emitted by the particles in the volume as the ray travels from $t{=}0$ to $D$, composited on top of a background color.
Since the probability density for stopping at $t$ is $\trans(t) \cdot \density(t)$, the expected color is
\begin{align}
\C = \int_0^D \trans(t) \cdot \density(t) \cdot \radiance(t) \; dt \;+\; \trans(D) \cdot \radiance_\text{bg}
\label{eq:volren}
\end{align}
where $\radiance_\text{bg}$ is a background color that is composited with the foreground scene according to the residual transmittance~$\trans(D)$.
Without loss of generality, we omit the background term in what follows.

\paragraph{Homogeneous media}
We can calculate the color of some homogeneous volumetric media with constant color $\radiance_a$ and density $\density_a$ over a ray segment $[a,b]$ by integration:
\begin{align}
\C(a \rightarrow b)
&= \int_a^b \trans(a\rightarrow t) \cdot \density(t) \cdot \radiance(t)  \; dt
\\
&= \density_a \cdot \radiance_a \int_a^b \trans(a\rightarrow t) \; dt
&\text{constant density/radiance}
\\
&= \density_a \cdot \radiance_a \int_a^b \exponential{-\int_a^t \density(u) \; du} \; dt
&\text{ substituting \eq{trans_ab}}
\\
&= \density_a \cdot \radiance_a \int_a^b \exponential{- \left. \density_a u \right|_a^t} \; dt
&\text{constant density (again)}
\\
&= \density_a \cdot \radiance_a \int_a^b \exponential{- \density_a (t-a)} \; dt \\
&= \density_a \cdot \radiance_a \left. \cdot \frac{\exponential{-\density_a (t-a)}}{-\density_a} \right|_a^b \\
&= \radiance_a \cdot (1 - \exponential{-\density_a (b-a)}) 
\label{eq:homogeneous}
\end{align}

\paragraph{Transmittance is multiplicative}
Note that transmittance factorizes as follows:
\begin{align}
\trans(a \rightarrow c) = &= \exponential{-\left[ \int_a^b \density(t) \; dt +  \int_b^c \density(t) \; dt \right]} 
\\
&= \exponential{- \int_a^b \density(t) \; dt } \exponential{ - \int_b^c \density(t) \; dt} 
\\
&= \trans(a \rightarrow b) \cdot \trans(b \rightarrow c)
\label{eq:factor}
\end{align}
This also follows from the probabilistic interpretation of $\trans$, since the probability that the ray does not hit any particles within $[a, c]$ is the product of the probabilities of the two independent events that it does not hit any particles within $[a, b]$ or within $[b, c]$.

\paragraph{Transmittance for piecewise constant data}
Given a set of intervals $\{[t_n, t_{n+1}]\}_{n=1}^N$ with constant density $\sigma_n$ within the $n$-th segment, and with $t_1{=}0$ and $\delta_n {=} t_{n+1} {-} t_n$, transmittance is equal to:
\begin{align}
\trans_n = \trans(t_n) = \trans(0 \rightarrow t_n) 
&= \exponential{- \int_{0}^{t_n} \density(t) \; dt} 
= \exponential{\sum_{k=1}^{n-1} -\density_k \delta_k}
\label{eq:trans_const}
\end{align}

\newpage
\paragraph{Volume rendering of piecewise constant data}
Combining the above, we can evaluate the volume rendering integral through a medium with piecewise constant color and density:
\begin{align}
\C(t_{N+1}) &= \sum_{n=1}^N \int_{t_n}^{t_{n+1}} \trans(t) \cdot \density_n \cdot \radiance_n \; dt
&\text{piecewise constant}
\\
&= \sum_{n=1}^N \int_{t_n}^{t_{n+1}} \trans(0 \rightarrow t_n) \cdot \trans(t_n \rightarrow t) \cdot \density_n \cdot \radiance_n \; dt &\text{from \eq{factor}}
\\
&= \sum_{n=1}^N \trans(0 \rightarrow t_n)  \int_{t_n}^{t_{n+1}} \trans(t_n \rightarrow t) \cdot \density_n \cdot \radiance_n \; dt &\text{constant}
\\
&= \sum_{n=1}^N \trans(0 \rightarrow t_n) \cdot (1 - \exponential{-\density_n (t_{n+1}-t_n)}) \cdot \radiance_n &\text{from \eq{homogeneous}}
\end{align}
This leads to the volume rendering equations from NeRF~\cite[Eq.3]{mildenhall2020nerf}:
\begin{align}
\C(t_{N+1}) = \sum_{n=1}^N \trans_n \cdot (1 - \exponential{-\density_n \delta_n}) \cdot \radiance_n 
\text{, \quad where \quad}
\trans_n = \exponential{\sum_{k=1}^{n-1} -\density_k \delta_k}
\label{eq:final_color}
\end{align}
Finally, rather than writing these expressions in terms of volumetric density, we can re-express them in terms of alpha-compositing weights $\weight_n \equiv 1-\exponential{-\density_n \delta_n}$, and by noting that $\prod_i \exp x_i = \exponential{\sum_i x_i}$ in \eq{trans_const}:
\begin{align}
\C(t_{N+1}) = \sum_{n=1}^N \trans_n \cdot \weight_n \cdot \radiance_n
\text{, \quad where \quad}
\trans_n = \prod_{n=1}^{N-1}(1-\weight_n)
\label{eq:final_color2}
\end{align}

\paragraph{Alternate derivation} By making use of the earlier connection between CDF and PDF that $(1 - \trans)' = \trans \sigma$, and by assuming constant color $\radiance_a$ along an interval $[a, b]$:
\begin{align}
\int_a^b \trans(t) \cdot \sigma(t) \cdot \radiance(t) \; dt &= \radiance_a \int_a^b (1-\trans)'(t) \; dt \\
&= \radiance_a \cdot  (1-\trans(t))  |_a^b \\
&= \radiance_a \cdot (\trans(a) - \trans(b)) \\
&= \radiance_a \cdot \trans(a)\cdot (1 - \trans(a\rightarrow b))
\end{align}
Combined with constant per-interval density, this identity yields the same expression for color as \eq{final_color}.

\bibliographystyle{plain}
\bibliography{main}

\section*{Acknowledgements}
Thanks to Daniel Rebain, Soroosh Yazdani and Rif A. Saurous for a careful proofread.
\end{document}

%% file: preamble.tex
\usepackage{color}
\usepackage{amsmath}
\usepackage{amssymb}
\usepackage{amsfonts}
\usepackage{url}

\definecolor{turquoise}{cmyk}{0.65,0,0.1,0.3}
\definecolor{purple}{rgb}{0.65,0,0.65}
\definecolor{dark_green}{rgb}{0, 0.3, 0}
\definecolor{orange}{rgb}{0.8, 0.6, 0.2}
\definecolor{red}{rgb}{0.8, 0.2, 0.2}
\definecolor{darkred}{rgb}{0.6, 0.1, 0.05}
\definecolor{blueish}{rgb}{0.0, 0.3, .6}
\definecolor{light_gray}{rgb}{0.7, 0.7, .7}
\definecolor{pink}{rgb}{1, 0, 1}
\definecolor{greyblue}{rgb}{0.25, 0.25, 1}
\definecolor{gold}{rgb}{0.7, 0.5, 0}

\renewcommand{\paragraph}[1]{\vspace{1em}\noindent\textbf{#1}.}
\newcommand{\eq}[1]{(\ref{eq:#1})}